\documentclass{article}
\usepackage[utf8]{inputenc}
\usepackage{graphicx}
\usepackage{parskip}

\pdfoutput=1
\usepackage[final]{cml4impact_2022}
\usepackage{hyperref}
\usepackage{amsmath}
\usepackage[capitalize,noabbrev]{cleveref}
\Crefname{equation}{Equation}{Equations}
\Crefname{figure}{Figure}{Figures}
\Crefname{tabular}{Table}{Tables}

\usepackage[utf8]{inputenc} 
\usepackage[T1]{fontenc}    
\usepackage{url}            
\usepackage{booktabs}       
\usepackage{amsfonts}       
\usepackage{nicefrac}       
\usepackage{microtype}      
\usepackage{xcolor}         
\usepackage{subcaption}

\title{Can Large Language Models Build Causal Graphs?}

\author{%
  Stephanie Long\thanks{stephanie.long@mail.mcgill.ca} \\
Dept. of Family Medicine, \\
McGill University\\
   \And
  Tibor Schuster\\
   Dept. of Family Medicine,\\
   McGill University \\
 \And
   Alexandre Pich\'{e} \\
   Mila, Universit\'{e} de Montr\'{e}al\\
   ServiceNow Research \\
}

\begin{document}

\maketitle

\begin{abstract}

Building causal graphs can be a laborious process. To ensure all relevant causal pathways have been captured, researchers often have to discuss with clinicians and experts while also reviewing extensive relevant medical literature. By encoding common and medical knowledge, large language models (LLMs) represent an opportunity to ease this process by automatically scoring edges (i.e., connections between two variables) in potential graphs. LLMs however have been shown to be brittle to the choice of probing words, context, and prompts that the user employs. In this work, we evaluate if LLMs can be a useful tool in complementing causal graph development.

\end{abstract}

\section{Introduction}

Advances in causal inference have important implications in empirical research as most research questions asked in the health and medical context are not associational, but \emph{causal} in nature. Examples of such research questions include: \emph{What is the efficacy of a given drug in a given population? What is the expected effect of a given intervention on a specific outcome?} Common amongst these research questions is the desire to uncover the cause-and-effect relationships amongst a set of variables i.e., treatments, interventions, and outcomes. Such \emph{causal} questions cannot be answered from (observed) data alone or from the distributions that govern said data \citep{pearl2009}. In addition, external knowledge is needed to understand the underlying data-generating mechanisms to enable the setup of an appropriate 'inference engine'.

Causal diagrams play a central role in causal inference because they encode contextual knowledge of the observable and unobservable variables, and their causal dependencies. Causal inference pioneer Judea Pearl refers to the nodes in a causal diagram as a “\emph{society of listening variables}” \citep{pearl2017}. The term “\emph{listening}” stresses the defining property of directed and acyclic relationships between the variables, i.e., listening being asymmetrical, variable A listening to variable B, does not imply variable B listening to variable A, motivating the commonly adapted nomenclature of Directed Acyclic Graphs (DAGs) \citep{greenland1999causal, greenland2007causal}.

The first step when aiming to address causal questions using data is to draw a causal diagram e.g., a causal DAG. However, with the growing complexity and depth of health and medical knowledge being generated and increasing availability of new research articles daily, research databases are reaching dimensions that limit the possibility of parsing through the enormity of evidence needed to craft comprehensive DAGs \citep{rag2014}. Though expert opinion is the most valuable tool for drawing DAGs, experts do not always generate perfect DAGs, sometimes missing important confounding pathways \citep{oates2017}. Additionally, obtaining the opinions of numerous experts is costly both in time and resources. Thus, the ongoing developments of Large Language Models (LLM) may offer promise to help overcome some of these challenges by leveraging existing text data that may express causal sentiments (e.g., "X causes Y"). 

This research aims to answer the question, "\textit{Can large language models help researchers build causal diagrams in the medical context using existing text data?}" Here we will conduct experiments to determine under what conditions (e.g., prompt engineering, use of alternative language) GPT-3 \citep{brown2020language} is able to provide accurate answers regarding the relationship between variables in a medical context and what are its limitations in doing so.

The main contributions of this paper are:
\begin{itemize}
    \item Determining whether GPT-3 can signal the presence or absence of an edge between two variables in a directed acyclic graph from the medical context. 
    \item Evaluating whether the use of certain language in prompts or linking verbs improves the classification accuracy of GPT-3.
    \item Exploring the limitations of GPT-3 in understanding the causal relationships between variables in the medical context. 
\end{itemize}

\begin{figure}[t]
    \centering
    \includegraphics[width=\textwidth]{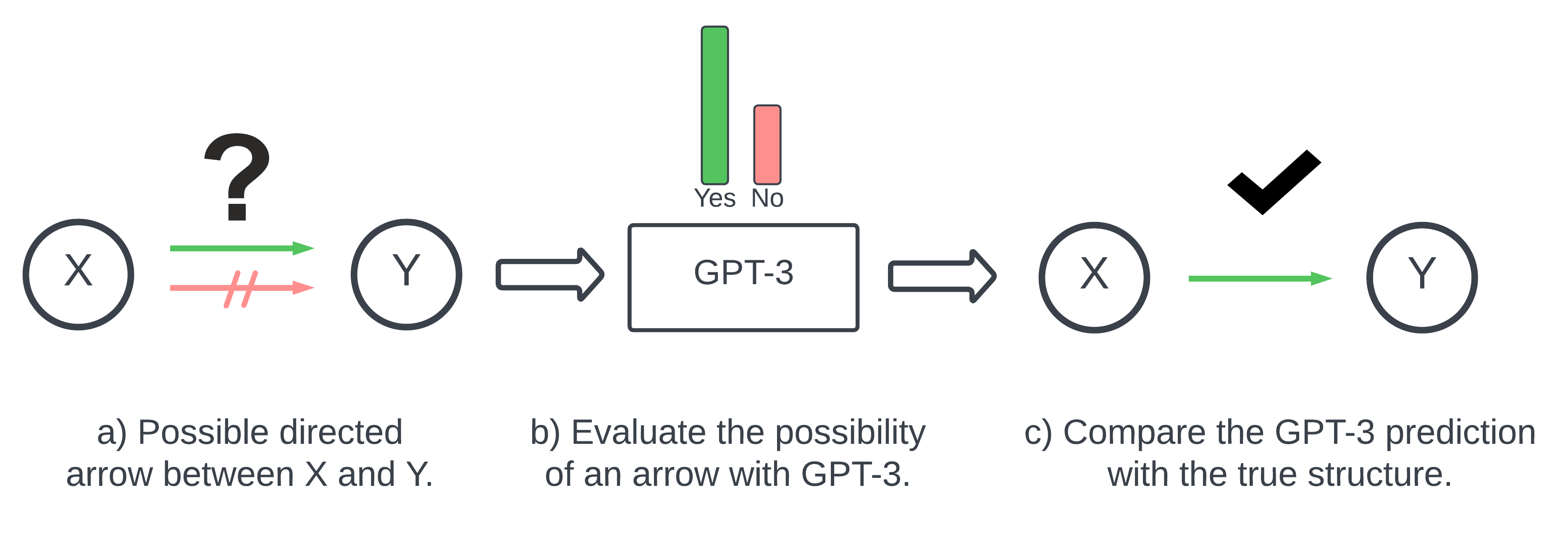}
    \caption{\textbf{Overview of the evaluation.} To predict the structure of a given causal graph, for every ordered variable pair, we scored two statements using GPT-3, where the first statement implied the presence of an arrow and the second implied the absence of an arrow. GTP-3 was accurate if the correct statement had a higher accuracy score than the incorrect statement. For example, GPT-3 would be accurate if the statement implying the presence (or absence) of an arrow had a higher accuracy than the incorrect statement and the arrow was present (or absent) in the true DAG.}
    \label{fig:overview}
\end{figure}

\section{Background}

\subsection{Large language models}


Large language models capture non-trivial relationships and knowledge about the datasets they have been trained upon. This knowledge has the possibility to unlock numerous applications in healthcare such as summarizing research papers, assessing patient risks from subjective symptoms, and diagnosing patients from clinical notes.

Although LLMs perform well on general natural language processing (NLP) tasks, its performance has been shown to be sensitive to its prompt \citep{moradi2021gpt, gutierrez2022thinking}. The advent of \textit{prompt-based learning} introduced a possible solution to context sensitive text, by querying LLMs with a prompt that uses in-domain examples or task descriptions \citep{liu2021pre}. For example, chain-of-thought prompts such as \textit{"Let's take this step by step"} have been shown to trigger multi-step reasoning in solving arithmetic problems \citep{kojima2022large}. Such prompts have also been shown to significantly improve performance in reasoning about medical questions \citep{lievin2022can}. 

Large language models are also sensitive to the type of text data they are trained on. For instance, GPT-3 \citep{brown2020language} was trained on the corpus of text information on the internet. As one can imagine, the entirety of the internet would include a range of text data from lay and casual use of language on social media to more formal language in news articles. These differences in writing styles may influence the frequency of the use of causal language describing non-causal relationships. For instance, an individual writing a social media post may use the word 'cause' more lightly than medical researchers in medical journals.

\subsection{Causal diagram overview}

Causal models are typically accompanied by graphical representations i.e., Directed Acyclic Graphs (DAGs) which are acyclic graphs that succinctly illustrate the qualitative assumptions made by the models, not captured by conventional statistical models or machine learning algorithms \citep{greenland2002overview, greenland1999causal}. 

In epidemiological research, DAGs have a variety of purposes including: (1) representing the causal relationships amongst variables \citep{greenland2002overview,greenland2007causal, pearl1995causal}; (2) identifying the potential confounding variables which need to be controlled for in order to estimate causal effects \citep{greenland2007causal, pearl1995causal, robins2001data, hernan2002causal}; and more recently (3) as a means of classifying the types of causal relationships that may give rise to selection bias \citep{hernan2004structural}. 

A DAG is composed of variables (nodes), both measured and unmeasured, and their connections are displayed via line segments (directed edges) \citep{greenland1999causal, hernan2004structural}. The \textit{absence} of an arrow between variables indicates the lack of a direct relationship between the variables. If the edge has an arrowhead, the variable at the tail is the parent node and the variable at the arrowhead is the child node \citep{greenland2007causal}. An edge or arc is any line (with an arrowhead or not) that connects two variables \citep{greenland2002overview}. The main characteristics of DAGs are that they are: (1) \textit{directed} i.e., the edge has a defined direction (arrowhead), and (2) \textit{acyclic} i.e., lack of cycles or loops within the graph.



\begin{figure}[t]
    \centering
    \includegraphics[width=\textwidth]{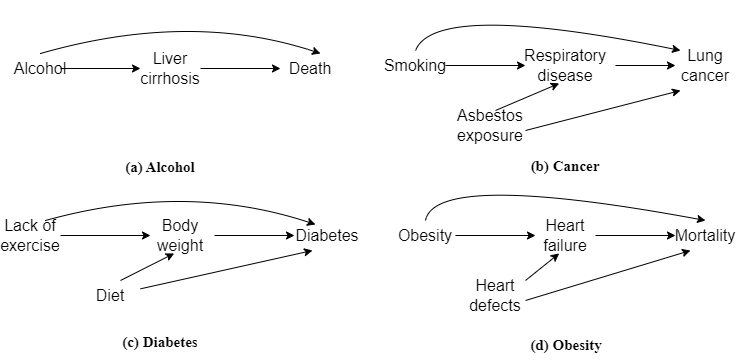} 
    \caption{\textbf{Ground Truth DAGs.} Four DAGs illustrating well-known exposure-outcome effects in the medical literature. DAG (A) represents the simplest DAG evaluated by GPT-3. DAGs (B-D) represent more complex structures involving a collider variable  (node with two arrows pointing into it e.g., 'respiratory disease', 'body weight', and 'heart failure') with a common cause with the outcome.}
    \label{fig:" diagram illustrating the relationship between obesity and mortality."}
\end{figure}

A DAG is causal if: (1) the arrows between variables can be interpreted as direct causal effects, and (2) all common causes of any pair of variables are present \citep{hernan2004structural}. The causal effects are ‘direct’ relative to certain degrees of abstraction in that the DAG does not include any variables that may mediate the effect \citep{greenland2007causal}. As the name suggests, DAGs are acyclic because a variable cannot be the cause of itself, either directly or indirectly through another variable i.e., there are no feedback loops; as illustrated by each DAG in Figure~\ref{fig:" diagram illustrating the relationship between obesity and mortality."} \citep{hernan2004structural}. Additionally, in DAGs, causal pathways are represented with directed paths from the starting variable to the final variable; thus, a variable is the cause of its descendants and an effect of its ancestors \citep{greenland2007causal}.


\section{Experiments}



\subsection{Experimental details}

To empirically assess the potential effectiveness of LLMs in building DAGs, we used four DAGs representing well-known exposure-outcome relationships in the medical literature (\cref{fig:" diagram illustrating the relationship between obesity and mortality."}) as the Ground Truth. These DAGs are varied in complexity, amount of variables, and reflect different medical contexts. For a DAG of \textit{N} variables, there are $\tbinom{N}{2}$ possible edges between two variables, and there are twice this amount of possible arrows since the arrows are directed. For example, a DAG of 4 variables has $2\times \tbinom{4}{2} = 12$ possible arrows.

For each DAG, we looped through every ordered variable pair, and asked GPT-3 to score two statements per pair: (1) one implying the presence of a directed edge from variable 1 to variable 2, and (2) one implying absence of a directed edge from variable 1 to variable 2. The presence or absence of an edge between two variables is a binary decision (Yes / No), thus, we defined the prediction as accurate, if GPT-3 scored the correct statement higher than the incorrect one. We reported the \emph{accuracy} or the proportion of correct predictions of our model.

\subsection{Results}


\paragraph{Q1: Does using prompt engineering lead to more accurate answers?} We investigated if the prediction accuracy of GPT-3 could be improved by prompting the statements with a reference to a medical authority. For example,

\begin{center}
    "\emph{According to X, {var1} increases the risk of developing {var2}}", 
    
    instead of
    
    "\emph{{Var1} increases the risk of developing {var2}}" (baseline),
\end{center}

where X is an individual or entity with medical authority or expertise, e.g., medical doctors, medical studies, or "Big Pharma". These prompts were chosen as they vary in their credibility with the public. We found that in 2 cases (Diabetes and Obesity DAGs) prompt engineering did not help and baseline (no prompting individual or authority) outperformed all other prompts. While in 2 other cases, the "According to medical doctors," prompting significantly improved the accuracy of GPT-3. Interestingly, conditioning on "According to Big Pharma," decreases the accuracy of 3 of the 4 DAGs compared to the baseline. Furthermore, prompting the model on medical studies or medical doctors resulted in different results for half the DAGs. See ~\cref{tab:prompt_engineering} for all results.


\begin{table}[h]
    \centering
\begin{tabular}{llr}
\toprule
\textbf{DAG name} & \textbf{Prompt} &  \textbf{Accuracy}          \\
\midrule
Alcohol & Baseline &      0.33 \\
        & Big Pharma &      0.50 \\
        & \textbf{Medical doctors} &      \textbf{0.83} \\
        & Medical studies &      0.67 \\
          \cmidrule{1-3}
Cancer & Baseline &      0.75 \\
        & Big Pharma &      0.58 \\
        & \textbf{Medical doctors} &      \textbf{1.00} \\
        & \textbf{Medical studies} &      \textbf{1.00} \\
          \cmidrule{1-3}
Diabetes & \textbf{Baseline} &      \textbf{0.67} \\
        & Big Pharma &      0.50 \\
        & Medical doctors &      0.33 \\
        & Medical studies &      0.42 \\
          \cmidrule{1-3}
Obesity & \textbf{Baseline} &      \textbf{0.75} \\
        & Big Pharma &      0.58 \\
        & \textbf{Medical doctors} &      \textbf{0.75} \\
        & \textbf{Medical studies} &      \textbf{0.75}\\
\bottomrule
\end{tabular}
    \caption{\textbf{Prompt engineering:} The medical authority used to prompt the statement.}
    \label{tab:prompt_engineering}
\end{table}

\paragraph{Q2: Does the verb used to denote the relationship between the variables have an impact on accuracy?} For instance, "Variable 1 \textit{X} Variable 2" where \textit{X} represents the verb (or phrase) that denotes the relationship between the variables, e.g., "causes" or "increases the risk". 

Our results demonstrated that while no verb consistently improved classification accuracy, the choice of verb linking the two variables of interest influenced accuracy. 'Increases risk' had the highest accuracy for three of the four DAGs. Though it did not achieve the highest accuracy in the Alcohol DAG. Overall, the use of 'cause' yielded decent results for all DAGs. Results are reported in ~\cref{tab:causation_verb}.


\begin{table}[ht]
    \centering
\begin{tabular}{llr}
\toprule
\textbf{DAG name} & \textbf{Linking Verb} &  \textbf{Accuracy}   \\
\midrule
Alcohol & Cause &      0.33 \\
        & \textbf{Increases likelihood}  &      \textbf{0.50} \\
        & Increases risk  &      0.33 \\
          \cmidrule{1-3}
Cancer & Cause &      0.58 \\
        & Increases likelihood  &      0.58 \\
        & \textbf{Increases risk}  &      \textbf{0.75} \\
          \cmidrule{1-3}
Diabetes & Cause &      0.58 \\
        & Increases likelihood  &      0.42 \\
        & \textbf{Increases risk}  &      \textbf{0.67} \\
          \cmidrule{1-3}
Obesity & Cause &      0.58 \\
        & Increases likelihood  &      0.42 \\
        & \textbf{Increases risk}  &      \textbf{0.75} \\
\bottomrule
\end{tabular}
    \caption{\textbf{Linking verb:} The verb or phrase used to link the two variables of interest. }
    \label{tab:causation_verb}
\end{table}


\paragraph{Q3: Does specificity in language improve accuracy?} We investigated if making our statements more specific or descriptive improved GPT-3's accuracy. 

Unsurprisingly, rephrasing the "alcohol" variable to "excessive alcohol consumption" increased the accuracy of GPT-3 on the Alcohol DAG. However, being more specific about the number of cigarettes being smoked and using a clinical term to qualify obesity resulted in worse accuracy for the Cancer and Obesity DAGs. Overall, In this analysis, more specific statements did not increase the accuracy and often resulted in worse accuracy for different linking verbs. Results are reported in~\cref{tab:specificity}.

\begin{table}[h]
    \centering
\begin{tabular}{lllr}
\toprule
\textbf{DAG name} & \textbf{Variable Name} & \textbf{Linking Verb} &        \textbf{Accuracy}  \\
\midrule
Alcohol & Alcohol & Cause &      0.33 \\
 &                            & Increases risk &      0.50 \\
          \cmidrule{2-4}
& \textbf{Excessive alcohol consumption} & Cause &      0.33 \\
&                              & \textbf{Increases risk} &      \textbf{0.67} \\
          \cmidrule{1-4}
Cancer & \textbf{Cigarette smoking} & Cause &      0.58 \\
&                              & \textbf{Increases risk} &      \textbf{0.67} \\
          \cmidrule{2-4}
& Smoking 100 cigarettes a day & Cause &      0.50 \\
&                              & Increases risk &      0.58 \\
          \cmidrule{1-4}
Obesity & \textbf{Obesity} & Cause &      0.58 \\
&                              & \textbf{Increases risk} &      \textbf{0.67} \\
          \cmidrule{2-4}
& Excessive fat accumulation & Cause &      0.58 \\
&                              & Increases Risk &      0.58 \\
\bottomrule
\end{tabular}
    \caption{\textbf{Specificity:} More extensive descriptions of variables/concepts.}
    \label{tab:specificity}
\end{table}


\section{Discussion}

In this work, we explored if LLMs could be used to complement and speed up the workflow of researchers by automatically scoring edges in potential DAGs. For the relatively simple and well-studied DAGs that we tested GPT-3 on, the results were overall encouraging as the performance reached much higher than 50\% accuracy (random guessing) on all DAGs for at least one of the tested settings (e.g., prompt or linking verb). In this analysis, we found that GPT-3's accuracy performance was influenced by different prompts and linking verbs between variables of interest. 

To the best of our knowledge, this is the first study to examine using LLM for causal diagram development in the medical context. Though there is growing interest, to date, there are few studies exploring the utility of LLM in causal diagram development. A recent study by \cite{willig2022can} compared the performance of three query LLMs in making causal graph predictions in a general context. There also has been some interesting works applying causal inference in the LLM context. For instance, \cite{vig2020investigating} investigated gender bias present in LLM using causal mediation analysis. \cite{feder2021causal} released a preprint of a consolidated exploration of causal inference situated in NLP. These works suggest more focus is being devoted to researching how causal inference can be applied to LLMs and NLP. 

Furthermore, there has been some research investigating LLM's ability to answer and reason with medical text data. Several recent studies \citep{lievin2022can, guo2022} showed promising results on LLMs ability to answer medical exam questions. Others \citep{moradi2021gpt, gutierrez2022thinking} have shown that context-specific LLMs such as BioBert are able to outperform GPT-3 in medical domain NLP tasks. 

\paragraph{Limitations} This study has some limitations. First, it must be acknowledged that the updating of LLMs, themselves as well as the data they are trained upon, lags behind the availability of new medical literature, and, thus may not be useful for informing the building of DAGs for novel diseases. Additionally, GPT-3 was trained upon the corpus of text data uploaded to the internet. The language used on the broader internet is likely more casual with the use of causal language than the medical academic literature \citep{haber2022causal}. Lastly, the way in which we probed GPT-3's ability to draw an edge between variables assumes that the causal connections between variables would be well-established in the corpus of text data.


\paragraph{Future work}
Future work aims to use a medical language context-specific LLM such as web-GPT with PubMed  or BioBert \citep{lee2020biobert} to signal the presence or absence of edges in DAGs using medical terminology. Additionally, since our preliminary evaluations only examined the presence/absence of arrows and their direction, upcoming projects will be focused on controlling for acyclicity amongst variables, another important characteristic of DAGs.

\section{Conclusion}
Our results illustrate that GPT-3's level of accuracy in confirming an edge connecting two variables in a DAG depends on the language used to describe the relationship. Presently, expert opinion is the most valuable tool for constructing DAGs; however, like LLMs, experts are not exempt from making errors resulting in imperfect or erroneous DAGs via omission of important confounder variables \citep{oates2017}. These imperfections highlight that the use of LLMs to build DAGs should be, at present, only conducted with expert verification. We see LLMs providing utility in extracting common knowledge from medical text which when paired with expert knowledge may present a more efficient means to generate comprehensive DAGs. 

Large Language Models represent an exciting opportunity to extract common knowledge from the medical literature to complement and speed up DAG creation, but further research must be done to address the limitations reported above.
%


\newpage

\bibliography{bibliography}

\begin{thebibliography}{23}
\providecommand{\natexlab}[1]{#1}
\providecommand{\url}[1]{\texttt{#1}}
\expandafter\ifx\csname urlstyle\endcsname\relax
  \providecommand{\doi}[1]{doi: #1}\else
  \providecommand{\doi}{doi: \begingroup \urlstyle{rm}\Url}\fi

\bibitem[Brown et~al.(2020)Brown, Mann, Ryder, Subbiah, Kaplan, Dhariwal,
  Neelakantan, Shyam, Sastry, Askell, et~al.]{brown2020language}
T.~Brown, B.~Mann, N.~Ryder, M.~Subbiah, J.~D. Kaplan, P.~Dhariwal,
  A.~Neelakantan, P.~Shyam, G.~Sastry, A.~Askell, et~al.
\newblock Language models are few-shot learners.
\newblock \emph{Advances in neural information processing systems},
  33:\penalty0 1877--1901, 2020.

\bibitem[Feder et~al.(2021)Feder, Keith, Manzoor, Pryzant, Sridhar,
  Wood-Doughty, Eisenstein, Grimmer, Reichart, Roberts,
  et~al.]{feder2021causal}
A.~Feder, K.~A. Keith, E.~Manzoor, R.~Pryzant, D.~Sridhar, Z.~Wood-Doughty,
  J.~Eisenstein, J.~Grimmer, R.~Reichart, M.~E. Roberts, et~al.
\newblock Causal inference in natural language processing: Estimation,
  prediction, interpretation and beyond.
\newblock \emph{arXiv preprint arXiv:2109.00725}, 2021.

\bibitem[Greenland and Brumback(2002)]{greenland2002overview}
S.~Greenland and B.~Brumback.
\newblock An overview of relations among causal modelling methods.
\newblock \emph{International journal of epidemiology}, 31\penalty0
  (5):\penalty0 1030--1037, 2002.

\bibitem[Greenland and Pearl(2006)]{greenland2007causal}
S.~Greenland and J.~Pearl.
\newblock Causal diagrams.
\newblock \emph{Encyclopedia of Epidemiology}, 2006.

\bibitem[Greenland et~al.(1999)Greenland, Pearl, and
  Robins]{greenland1999causal}
S.~Greenland, J.~Pearl, and J.~M. Robins.
\newblock Causal diagrams for epidemiologic research.
\newblock \emph{Epidemiology}, pages 37--48, 1999.

\bibitem[Guo et~al.(2022)Guo, Cao, and Yi]{guo2022}
Q.~Guo, S.~Cao, and Z.~Yi.
\newblock A medical question answering system using large language models and
  knowledge graphs.
\newblock \emph{International Journal of Intelligent Systems}, 37\penalty0
  (11):\penalty0 8548--8564, 2022.
\newblock \doi{https://doi.org/10.1002/int.22955}.

\bibitem[Guti{\'e}rrez et~al.(2022)Guti{\'e}rrez, McNeal, Washington, Chen, Li,
  Sun, and Su]{gutierrez2022thinking}
B.~J. Guti{\'e}rrez, N.~McNeal, C.~Washington, Y.~Chen, L.~Li, H.~Sun, and
  Y.~Su.
\newblock Thinking about gpt-3 in-context learning for biomedical ie? think
  again.
\newblock \emph{arXiv preprint arXiv:2203.08410}, 2022.

\bibitem[Haber et~al.(2022)Haber, Wieten, Rohrer, Arah, Tennant, Stuart,
  Murray, Pilleron, Lam, Riederer, et~al.]{haber2022causal}
N.~Haber, S.~Wieten, J.~Rohrer, O.~Arah, P.~Tennant, E.~Stuart, E.~Murray,
  S.~Pilleron, S.~Lam, E.~Riederer, et~al.
\newblock Causal and associational language in observational health research: a
  systematic evaluation.
\newblock \emph{American Journal of Epidemiology}, 2022.

\bibitem[Hern{\'a}n et~al.(2002)Hern{\'a}n, Hern{\'a}ndez-D{\'\i}az, Werler,
  and Mitchell]{hernan2002causal}
M.~A. Hern{\'a}n, S.~Hern{\'a}ndez-D{\'\i}az, M.~M. Werler, and A.~A. Mitchell.
\newblock Causal knowledge as a prerequisite for confounding evaluation: an
  application to birth defects epidemiology.
\newblock \emph{American journal of epidemiology}, 155\penalty0 (2):\penalty0
  176--184, 2002.

\bibitem[Hern{\'a}n et~al.(2004)Hern{\'a}n, Hern{\'a}ndez-D{\'\i}az, and
  Robins]{hernan2004structural}
M.~A. Hern{\'a}n, S.~Hern{\'a}ndez-D{\'\i}az, and J.~M. Robins.
\newblock A structural approach to selection bias.
\newblock \emph{Epidemiology}, pages 615--625, 2004.

\bibitem[Kojima et~al.(2022)Kojima, Gu, Reid, Matsuo, and
  Iwasawa]{kojima2022large}
T.~Kojima, S.~S. Gu, M.~Reid, Y.~Matsuo, and Y.~Iwasawa.
\newblock Large language models are zero-shot reasoners.
\newblock \emph{arXiv preprint arXiv:2205.11916}, 2022.

\bibitem[Lee et~al.(2020)Lee, Yoon, Kim, Kim, Kim, So, and
  Kang]{lee2020biobert}
J.~Lee, W.~Yoon, S.~Kim, D.~Kim, S.~Kim, C.~H. So, and J.~Kang.
\newblock Biobert: a pre-trained biomedical language representation model for
  biomedical text mining.
\newblock \emph{Bioinformatics}, 36\penalty0 (4):\penalty0 1234--1240, 2020.

\bibitem[Li{\'e}vin et~al.(2022)Li{\'e}vin, Hother, and Winther]{lievin2022can}
V.~Li{\'e}vin, C.~E. Hother, and O.~Winther.
\newblock Can large language models reason about medical questions?
\newblock \emph{arXiv preprint arXiv:2207.08143}, 2022.

\bibitem[Liu et~al.(2021)Liu, Yuan, Fu, Jiang, Hayashi, and Neubig]{liu2021pre}
P.~Liu, W.~Yuan, J.~Fu, Z.~Jiang, H.~Hayashi, and G.~Neubig.
\newblock Pre-train, prompt, and predict: A systematic survey of prompting
  methods in natural language processing.
\newblock \emph{arXiv preprint arXiv:2107.13586}, 2021.

\bibitem[Moradi et~al.(2021)Moradi, Blagec, Haberl, and Samwald]{moradi2021gpt}
M.~Moradi, K.~Blagec, F.~Haberl, and M.~Samwald.
\newblock Gpt-3 models are poor few-shot learners in the biomedical domain.
\newblock \emph{arXiv preprint arXiv:2109.02555}, 2021.

\bibitem[Oates et~al.(2017)Oates, Kasza, Simpson, and Forbes]{oates2017}
C.~Oates, J.~Kasza, J.~Simpson, and A.~Forbes.
\newblock Repair of partly misspecified causal diagrams.
\newblock \emph{Epidemiology}, 28, 2017.

\bibitem[Pearl(1995)]{pearl1995causal}
J.~Pearl.
\newblock Causal diagrams for empirical research.
\newblock \emph{Biometrika}, 82\penalty0 (4):\penalty0 669--688, 1995.

\bibitem[Pearl(2009)]{pearl2009}
J.~Pearl.
\newblock Causal inference in statistics: An overview.
\newblock \emph{Statistics Surveys}, 2009.

\bibitem[Pearl(2017)]{pearl2017}
J.~Pearl.
\newblock The eight pillars of causal wisdom.
\newblock \emph{UCLA}, 2017.

\bibitem[Raghupathi(2014)]{rag2014}
V.~Raghupathi, W.;~Raghupathi.
\newblock Big data analytics in healthcare- promise and potential.
\newblock \emph{Health Information Science and Systems}, 2, 2014.

\bibitem[Robins(2001)]{robins2001data}
J.~M. Robins.
\newblock Data, design, and background knowledge in etiologic inference.
\newblock \emph{Epidemiology}, pages 313--320, 2001.

\bibitem[Vig et~al.(2020)Vig, Gehrmann, Belinkov, Qian, Nevo, Singer, and
  Shieber]{vig2020investigating}
J.~Vig, S.~Gehrmann, Y.~Belinkov, S.~Qian, D.~Nevo, Y.~Singer, and S.~Shieber.
\newblock Investigating gender bias in language models using causal mediation
  analysis.
\newblock \emph{Advances in Neural Information Processing Systems},
  33:\penalty0 12388--12401, 2020.

\bibitem[Willig et~al.(2022)Willig, Ze{\v{c}}evi{\'c}, Dhami, and
  Kersting]{willig2022can}
M.~Willig, M.~Ze{\v{c}}evi{\'c}, D.~S. Dhami, and K.~Kersting.
\newblock Can foundation models talk causality?
\newblock \emph{arXiv preprint arXiv:2206.10591}, 2022.

\end{thebibliography}
\bibliographystyle{abbrvnat}

\medskip

\appendix

\end{document}